\documentclass[runningheads]{llncs}
\usepackage{graphicx}
\usepackage{algorithm}
\usepackage{algpseudocode}
\usepackage[acronym]{glossaries}
\usepackage{subfig}
\usepackage{multirow}
\usepackage{multicol}
\usepackage[table]{xcolor}
\usepackage{bm}
\usepackage[hidelinks,bookmarks=true]{hyperref}
\usepackage[T1]{fontenc}
\usepackage{lmodern}
\usepackage{svg}

\newacronym{ai}{AI}{Artificial Intelligence}
\newacronym[plural=ANNs,firstplural=Artificial Neural Networks (ANNs)]{ann}{ANN}{Artificial Neural Network}

\newacronym{cfg}{CFG}{Context-free Grammar}
\newacronym[plural=CPUs,firstplural=Central Processing Units (CPUs)]{cpu}{CPU}{Central Processing Unit}
\newacronym[plural=CNNs,firstplural=Convolutional Neural Networks (CNNs)]{cnn}{CNN}{Convolutional Neural Network}

\newacronym[plural=DNNs,firstplural=Deep Neural Networks (DNNs)]{dnn}{DNN}{Deep Neural Network}
\newacronym{denser}{DENSER}{Deep Evolutionary Network Structured Evolution}
\newacronym{dsge}{DSGE}{Dynamic Structured Grammatical Evolution}

\newacronym{ea}{EA}{Evolutionary Algorithm}
\newacronym{ec}{EC}{Evolutionary Computation}
\newacronym[plural=ES,firstplural=Evolution Strategies (ES)]{es}{ES}{Evolution Strategy}

\newacronym{lemonade}{LEMONADE}{Lemarckian Evolutionary Algorithm for Multi-Objective Neural Architecture Design}

\newacronym{fdenser}{Fast-DENSER}{Fast Deep Evolutionary Network Structured Representation}
\newacronym{flops}{FLOPs}{floating point operations per second}

\newacronym{ge}{GE}{Grammatical Evolution}
\newacronym{gp}{GP}{Genetic Programming}
\newacronym[plural=GPUs,firstplural=Graphics Processing Units (GPUs)]{gpu}{GPU}{Graphics Processing Unit}

\newacronym{iot}{IoT}{Internet of Things}

\newacronym{ml}{ML}{Machine Learning}
\newacronym{moo}{MOO}{Multi-Objective Optimization}
\newacronym{mol}{MOL}{Multi-Output Learning}
\newacronym{monas}{MONAS}{Multi-Objective Neural Architecture Search}
\newacronym{mac}{MAC}{Multiply–accumulate operation}

\newacronym{nas}{NAS}{Neural Architecture Search}
\newacronym{ne}{NE}{Neuroevolution}
\newacronym{nsga-ii}{NSGA-II}{Nondominated Sorting Genetic Algorithm II}

\newacronym{rnn}{RNN}{Recurrent Neural Network}

\newacronym{sge}{SGE}{Structured Grammatical Evolution}

\begin{document}
\hypersetup{
	pdftitle={Towards Physical Plausibility in Neuroevolution Systems},
	pdfsubject={},
	pdfauthor={Gabriel Cortês,Nuno Lourenço,Penousal Machado},
	pdfkeywords={Evolutionary Computation,Neuroevolution,Energy Efficiency}
}
\title{Towards Physical Plausibility in Neuroevolution Systems}

\author{Gabriel Cortês\orcidID{0000-0001-6318-8520} \and
Nuno Lourenço\orcidID{0000-0002-2154-0642} \and
Penousal Machado\orcidID{0000-0002-6308-6484}}

\authorrunning{G. Cortês et al.}

\institute{University of Coimbra, CISUC/LASI -- Centre for Informatics and Systems of the University of Coimbra, Department of Informatics Engineering\\
\email{\{cortes,naml,machado\}@dei.uc.pt}}
\maketitle              

\begin{abstract}
The increasing usage of \gls{ai} models, especially \glspl{dnn}, is increasing the power consumption during training and inference, posing environmental concerns and driving the need for more energy-efficient algorithms and hardware solutions. This work addresses the growing energy consumption problem in \gls{ml}, particularly during the inference phase. Even a slight reduction in power usage can lead to significant energy savings, benefiting users, companies, and the environment. Our approach focuses on maximizing the accuracy of \gls{ann} models using a neuroevolutionary framework whilst minimizing their power consumption. To do so, power consumption is considered in the fitness function. We introduce a new mutation strategy that stochastically reintroduces modules of layers, with power-efficient modules having a higher chance of being chosen. We introduce a novel technique that allows training two separate models in a single training step whilst promoting one of them to be more power efficient than the other while maintaining similar accuracy. The results demonstrate a reduction in power consumption of \gls{ann} models by up to 29.2\% without a significant decrease in predictive performance.

\end{abstract}

\glsresetall

\section{Introduction}
\label{sec:introductiom}
As the demand for \gls{ml} continues to grow, so does the electrical power required for training and assessment. According to Patterson et al., GPT-3, the model behind ChatGPT, consumes 1287 MWh, corresponding to approximately 552 tons of CO$_2$ equivalent emissions just for training during 15 days \cite{patterson2022carbon}. In addition to the environmental impacts of this power usage, it can also burden individual users and organizations, who may face high energy costs. Therefore, finding ways to reduce the power consumption of \gls{ml} processes is becoming increasingly important.

\glspl{ann} are a type of \gls{ml} model inspired by biological neural networks \cite{yegnanarayana2009artificial}. They consist of multiple layers of artificial neurons, which are functions that take input data and produce an output based on it. The connections between neurons have an associated weight value modified in the training process to allow the network to ``learn'' how to solve a specific task. \glspl{dnn} are \glspl{ann} with a considerable number of hidden layers \cite{LecunDeepLearning,LiuDNNSurvey}. This allows them to avoid the feature engineering step, thus automatically discovering the representations needed for classification and achieving higher accuracy values. Training and executing \glspl{ann} is power-intensive due to the required computational resources.

\glspl{ea} are algorithms inspired by natural selection \cite{EibenSmithIntroEvoComputing,vikhar_ea_2016}. To evolve solutions over multiple generations, they utilize mechanisms, such as selection, crossover, and mutation. The process begins with a randomly initialized population whose evolution is steered by a fitness function that measures the quality of an individual. In conjunction with the mentioned evolutionary mechanisms, the process is predicted to culminate in near-optimal individuals.

\gls{ne} uses \glspl{ea} to generate and optimize \glspl{ann} for a given task \cite{GalvanNeuroEvolution}. It can optimize the \gls{ann}'s architecture and hyperparameters.

We hypothesise that we can address the energy inefficiency issue by using \gls{ne} to search for well-suited models for a particular problem while being power-efficient. \gls{fdenser} is a method that utilizes an \gls{es} to find optimal \gls{ann} models by using their accuracy as the fitness function, thus guiding the search towards accurate models \cite{AssuncaoFastDENSER}.

In this work, we propose novel approaches integrated into \gls{fdenser} to find power-efficient models. We have incorporated a new approach to measure the power consumption of a \gls{dnn} model during the inference phase. This metric has been embedded into multi-objective fitness functions to steer the evolution towards more power-efficient \gls{dnn} models. We also introduce a new mutation strategy that allows the reutilization of modules of layers with inverse probability to the power usage of a module, thus (re)introducing efficient sets of layers in a model. We propose the introduction of an additional output layer connected to an intermediate layer of a \gls{dnn} model and posterior partitioning into two separate models to obtain smaller but similarly accurate models that utilize less power. To the best of our knowledge, no prior works employ a similar approach.

The experiments are analyzed through two metrics: accuracy and mean power usage during the validation step. The motive for using the power usage of the validation step instead of the training step is that the training is usually performed only once. Contrarily, the inference is executed multiple times. Moreover, inference does not necessarily occur on the machine where the training was conducted, which is vital since many devices are not optimized for these tasks.

The results of this work show that it is possible to have \gls{dnn} models with substantially inferior power usage. The best model found regarding power consumes 29.18 W (29.2\%) less whilst having a tiny decrease in performance (less than 1\%).

This work is structured as follows: Section \ref{sec:background} provides background information on \glspl{ann}, and \gls{ne}. Section \ref{sec:approach} introduces our methodologies to enhance the power efficiency of \gls{ann} models. Section \ref{sec:experimental_setup} outlines the experimental setup. Section \ref{sec:results} presents the experimental results. Finally, in Section \ref{sec:conclusion}, we provide our conclusions and prospects for future research.

\section{Background}
\label{sec:background}

\subsection{Artificial Neural Networks}
\label{sec:background:ann}
Artificial Neural Networks are a type of supervised \gls{ml} inspired by biologic neural networks \cite{yegnanarayana2009artificial}. An \gls{ann} consists of connected processing units known as neurons. The connections follow a specific topology to achieve the desired application. A neuron's input may be the output of other neurons, external sources, or itself. Every connection has an associated weight, allowing the system to simulate biological synapses. A weighted sum of the inputs is computed at a given instant, considering the connection weights. It is also possible to sum a bias value to this. An activation function is applied, and thus, the neuron's output is obtained. 

\glspl{dnn} are \glspl{ann} composed of many hidden layers. Due to this, \glspl{dnn} can avoid the feature engineering step -- which usually requires human expertise -- by automatically discovering the representations needed for classification \cite{LecunDeepLearning,LiuDNNSurvey}. Thus, they can model more complex relationships and achieve higher accuracy on tasks requiring pattern recognition. The development and usage of \glspl{dnn} have substantially increased due to the widespread deployment of more capable hardware, such as \glspl{gpu} \cite{BalasDLHandbook}.

\subsection{Neuroevolution}
\label{sec:background:ne}
\gls{ne} is the application of evolutionary techniques to search for \gls{dnn} models. It is used to optimize the structure and weights of \glspl{dnn} to improve their performance on specific tasks, such as image classification and natural language processing. \gls{ne} is a gradient-free method based on the concept of population \cite{GalvanNeuroEvolution}. It allows for the simultaneous exploration of multiple zones of the search space through parallelization techniques at the cost of taking a usually long time to execute since each individual of the population is a \gls{dnn} that requires training and testing.

\gls{denser} is a neuroevolutionary framework that allows the search of \glspl{dnn} through a grammar-based neuroevolutionary approach that searches both network topology and hyperparameters \cite{AssuncaoDENSER}.

The developed \glspl{dnn} are structured according to a provided context-free grammar. \gls{denser} uses \gls{dsge} as the strategy that allows the modification of the network topology. \gls{dsge} is built upon \gls{sge}, with the main differences of allowing the growth of the genotype and only storing encoded genes \cite{dsge}. Allied with dynamic production rules, \gls{dsge} allows the creation of multiple-layer \glspl{dnn}. \gls{sge} proves to perform better than \gls{ge}, and \gls{dsge} proves to be superior to \gls{sge} \cite{sge}. The individuals of the evolutionary process are represented in two levels: the outer level encodes the topology of the \gls{ann}, and the inner one encodes its hyperparameters.

\gls{fdenser} was developed to overcome some limitations verified on \linebreak \gls{denser}: evaluating the population consumes a considerable amount of time, and the developed \glspl{dnn} are not fully trained \cite{AssuncaoFastDENSER}. \gls{fdenser} is an extension of DENSER on which the evolutionary engine is replaced by a $(1+\lambda)$-\gls{es}. This modification dramatically reduces the required number of evaluations per generation, enabling executions 20 times faster than the original version of \gls{denser}.

Moreover, individuals are initialized with shallow topologies, and the stopping criterion is variable to allow an individual to be trained for a more extended time.

On the CIFAR-10 dataset \cite{krizhevsky2009learning}, \gls{denser} obtained models with an accuracy higher than most of the state-of-the-art results, and on the CIFAR-100 \cite{krizhevsky2009learning}, it obtained the best accuracy reported by \gls{ne} approaches. \gls{fdenser} proves to be highly competitive relative to \gls{denser}, achieving execution times far inferior to its predecessor.
Additionally, \gls{fdenser} can develop \glspl{dnn} that do not require additional training after the evolutionary approach and are, therefore, ready to be deployed.

\section{Approach}
\label{sec:approach}

This section outlines the approaches developed to address the challenge of reducing power consumption in \gls{ann} models.

\subsection{Power Measurement}
\label{sec:approach:power}

Measuring the power a \gls{gpu} consumes is fundamental when developing approaches that minimize a model's energetic footprint. The ecosystem of developing a \gls{dnn} model mainly consists of three phases: design, training, and deployment. 

The design phase uses some energy, be it with manual design techniques or automatic methods. \gls{denser} is a \gls{ne} framework and, as such, consumes energy in the search for optimal models, and such consumption might be on par with the energy used on manual, trial-and-error methods. Reducing the energy used in this phase is out of the scope of this work.

The training of a \gls{dnn} model is an expensive process in which a model is trained on a large dataset to learn to predict unseen instances, taking a significant toll on technological companies' and individuals' power bills. While diminishing energy consumption during the training process remains a significant objective, it is worth noting that the inference phase in \glspl{dnn} holds vital importance during software deployment, as the software obtains results through inference. This becomes particularly relevant when considering the potential utilization of these models by millions of users. As such, tackling the minimization of energy consumption in this step is vital. For example, it is estimated that 80\% to 90\% of NVIDIA's \gls{ml} computations are inference processing \cite{luccioni2022estimating} and about 60\% of Google's \gls{ml} energy usage is for inference with the remaining portion being for training \cite{patterson2022carbon}.

Considering this, our work focuses on the power consumption in the inference step to allow a large deployment, thus saving more computational resources and energy and, on another layer, reducing financial expenses and reducing environmental impact.

\subsection{Model Partitioning}
\label{sec:approach:model_partitioning}
Training a \gls{dnn} model requires a substantial amount of time and considerable energy. Creating a process on which a single model is trained but can be split posteriorly into two models would reduce the time spent on training two models by, at most, two times. Pushing one of those two models into being smaller than the other may produce a simpler, similarly accurate, yet more power-efficient model.

\begin{figure}[hp]
	\centering
	\subfloat[Full model]{
		\includesvg[height=0.6\textwidth]{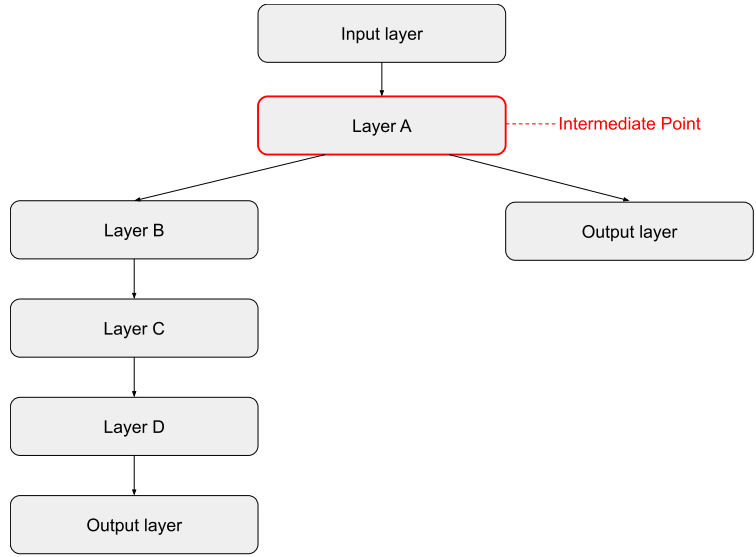}
		\label{fig:modelpartition:all}
	} \\
	\subfloat[Left partition]{
		\includesvg[width=0.27\textwidth]{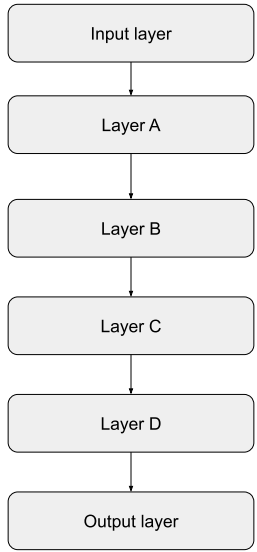}
		\label{fig:modelpartition:left}
	}
	\hspace{9.3em}
	\subfloat[Right partition]{
		\includesvg[width=0.27\textwidth]{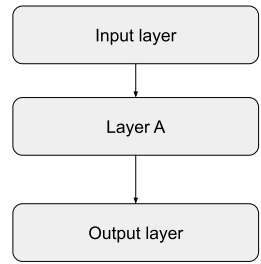}
		\label{fig:modelpartition:right}
	}
	\caption{Example of a two-output model and its left and right partitions, with the layer marked by the intermediate point in red.}
	\label{fig:modelpartition}
\end{figure}

Following this line of reasoning, we propose a modification to \gls{fdenser} on which an extra output layer is connected to an intermediate layer of the model. The two-output model (Figure \ref{fig:modelpartition:all}) is trained to optimize for two outputs. At the validation step, it is split into left (Figure \ref{fig:modelpartition:left}) and right (Figure \ref{fig:modelpartition:right}) partitions. These partitions are disjoint and can be evaluated similarly to how the complete model is evaluated, and metrics such as accuracy and power consumption can be obtained.

The intermediate point is a marker for where the additional output is added at the model partitioning step. We can, for example, consider a model as an array of layers, and the mentioned marker is the index of the layer to which the additional output is connected. This point can be assigned to any intermediate layer of the model. The input and output layers are excluded to prevent useless and redundant partitions. 

Since the maximum allowed value of the point is equal to the number of layers of the model minus one, the grammar initializer -- which generates individuals according to the grammar -- and the mutation mechanism for the macrostructure level of \gls{denser} -- which performs mutations on the hyperparameters of the individuals -- were modified to consider the maximum number of layers of the model dynamically. To introduce the intermediate point in the evolutionary process, it was considered part of the macrostructure and, as such, as a rule of the grammar. The introduced rule is \emph{<middle\_point> ::= [middle\_point,int,1,0,x]}, meaning that one integer value is obtained with the lower limit being zero. The upper limit is an arbitrary variable \emph{x} that is replaced at any instance by the maximum number of layers of the model minus one.

\subsection{Fitness Functions}
\label{sec:approach:mofitness}
To consider accuracy and power consumption in the fitness function, some functions were developed to take these parameters into account. Since our objective is to maximize accuracy but minimize power consumption, we consider the inverse of the latter, i.e., $power^{-1}$. 

Considering our approach of the division of a \gls{dnn} model into two comparably accurate partitions, with one smaller than the other, all of the presented fitness functions consider the accuracy of both partitions, intending to enhance both. These fitness functions only focus on minimizing power consumption within the larger partition, which is anticipated to experience higher power usage.

Firstly, as presented in Equation \ref{eq:f1}, we developed a fitness function that sums the accuracy of both partitions with the inverse of the power usage of the left partition. The accuracy values have an upper limit, consisting of minimum satisfiable values for the models, i.e., values below the state-of-the-art \cite{MeshkiniCNNFashionMNIST} to allow some tradeoff between accuracy and power consumption. The upper limit is higher on the right partition (0.85) than on the left partition (0.80) since it is desired that the right partition obtains a higher accuracy value, if possible. The goal of this function design was to obtain satisfiable models and, after that, guide the evolutionary process only by their power usage to minimize the power usage of the models. After testing, we observed that the power usage typically falls within the range $[30, 100]$ W, which, when inverted, resulted in values too small to be able to properly steer the evolutionary process.

\begin{equation}
	f_1 = \min(0.80, acc_{left}) + \min(0.85, acc_{right}) + power_{left}^{-1}
	\label{eq:f1}
\end{equation}

Considering this, another fitness function was designed (Equation \ref{eq:f2}), where the power usage is multiplied by 10, thus giving it a more considerable weight since power usage values for the used \gls{gpu} typically fall within the $[30, 100]$ W range. This weight is closely related to the used \gls{gpu} and should be modified accordingly. Preliminary experiments showed that although the evolution managed to somewhat minimize the power usage of the models, their accuracy remained around the chosen upper limits. Since this is not an optimal behaviour, a function that does not limit accuracy was developed.

\begin{equation}
	f_2 = \min(0.80, acc_{left}) + \min(0.85, acc_{right}) + 10 * power_{left}^{-1}
	\label{eq:f2}
\end{equation}

As shown in Equation \ref{eq:f3}, this fitness function considers only the accuracy of the partitions when both are below a threshold. After any of them surpass their respective threshold, power consumption is also considered, with a weight of 10. This means that, at first, evolution is only steered by the accuracy of the models. When satisfiable models are obtained, power consumption starts being considered to evolve both accurate and energy-efficient models. 

\begin{equation}
	f_3= 
	\begin{cases}
		acc_{left} + acc_{right} & \text{if } acc_{left} \leq 0.80 \land acc_{right} \leq 0.85 \\
		acc_{left} + acc_{right} + 10 * power_{left}^{-1},              & \text{otherwise}
	\end{cases}
	\label{eq:f3}
\end{equation}

\subsection{Module Reutilization}
\label{sec:approach:module_reuse}
Internally, \gls{fdenser} considers modules of layers on each individual from which a \gls{dnn} is then unravelled. One way to encourage the evolution of energy-efficient models is to provide an individual with a set of layers that are known to be efficient. As such, a scheme of module reutilization is proposed through the design of new mutation operators and the addition of an archive of modules and their respective power consumption.

Since this strategy only considers power consumption, it is expected that inaccurate models may sometimes be generated. Due to the nature of the evolutionary process and the used fitness function (Equation \ref{eq:f3}), inaccurate models are intensely penalized and, as such, discarded in favour of better ones.

Whenever a module of layers is randomly generated or modified, its power consumption is measured. To do this, a temporary model is created, which consists of an input layer, the module's layers, and an output layer. Since the module's accuracy is irrelevant, this temporary network is neither trained nor fed with a proper dataset, i.e., it is given random values instead of a dataset.

An operator of mutation, \emph{reuse module}, was introduced to take advantage of this information. It selects a module with a probability inversely proportional to its power consumption, i.e., modules with inferior power consumption have a superior probability of being chosen. As shown in Equation \ref{eq:p_i}, to obtain the probability of a module $i$ being chosen, we divide the inverse of its power, $power_i$, by the sum of the inverse power of all modules, with $n$ the number of saved modules. The selected module is introduced in a randomly chosen position. An operator that randomly removes a module from an individual is also introduced to counteract the described operator.

\begin{equation}
	P(i) = \frac{\frac{1}{power_i}}{\sum_{j=0}^{n}\frac{1}{power_j}}
	\label{eq:p_i}
\end{equation}

\section{Experimental Setup}
\label{sec:experimental_setup}
We performed two experiments: the baseline, which uses the plain version of \gls{fdenser} with accuracy as the fitness function, and an experiment where our proposed approaches were applied, using the fitness function presented in Equation \ref{eq:f3}. Table \ref{table:experimental:params} presents the experimental parameters used across the experiments. Note that \emph{DSGE-level rate} refers to the probability of a grammar mutation on the model's layers, the \emph{Macro layer rate} pertains to the probability of a grammar mutation affecting the macrostructure, encompassing elements such as hyperparameters or intermediate point mutation, and the \emph{Train longer rate} is the probability of allocating more time for an individual to be trained. The rates of reusing and removing modules do not apply to the baseline experiment. The experimental analyses consider the Mean Best Fitness (MBF) over 5 runs. 

The experiments were performed on a server running Ubuntu 20.04.3 LTS with an Intel Core i7-5930K CPU with a clock frequency of 3.50GHz, 32 GB of RAM, and an NVIDIA TITAN Xp with CUDA 11.2, CuDNN 8.1.0, Python 3.10.9, Tensorflow 2.9.1 and Keras 2.9.0 installed as well as the pyJoules 0.5.1 Python module with the NVIDIA specialization. 

\begin{table}[ht]
	\centering
	\caption{Experimental parameters}
	\label{table:experimental:params}
	\begin{tabular}{c|c}
		\textbf{Evolutionary Parameter} & \textbf{Value}            \\ \hline
		Number of runs                  & 5                         \\
		Number of generations           & 150                       \\
		Maximum number of epochs        & 10 000 000                \\
		Population size                 & 5                         \\
		Add layer rate                  & 25\%                      \\
		Reuse layer rate                & 15\%                      \\
		Remove layer rate               & 25\%                      \\
		Reuse module rate                & 15\%                      \\
		Remove module rate               & 25\%                      \\
		Add connection                  & 0\%                       \\
		Remove connection               & 0\%                       \\
		DSGE-level rate                 & 15\%                      \\
		Macro layer rate                & 30\%                      \\
		Train longer rate               & 20\%                      \\
		&                           \\
		\textbf{Train Parameter}        & \textbf{Value}            \\ \hline
		Default train time              & 10 min                    \\
		Loss                            & Categorical Cross-entropy
	\end{tabular}
\end{table}

All experiments used the Fashion-MNIST dataset \cite{FashionMNIST}, which was developed as a more challenging replacement for the well-known MNIST dataset \cite{deng2012mnist} by swapping handwritten digits with images of clothes such as shirts and coats, aiming at a more realistic and relevant benchmark. It is a balanced dataset consisting of a collection of 60 thousand examples for training and 10 thousand for testing, where each example is a 28x28 grey-scale image representing clothing items belonging to one of ten classes.

Since power usage is essential in making \gls{ne} physically plausible, a function to measure power was developed using the \emph{pyJoules} library. Its pseudocode can be analyzed in Algorithm \ref{alg:powerMeasure}, with \emph{meter} being the library tool that facilitates the measurement of energy consumed, and \emph{start} and \emph{stop} the functions that allow controlling it. It wraps a function call (\emph{func}, with corresponding arguments \emph{args}) while measuring the \gls{gpu} energetic consumption during its execution and the call's duration. This measurement is converted from milliJoule to Watt and appended to the array of measures. These steps are performed \emph{$n\_measures$} times, and then the mean value is calculated. In our work, we considered $n\_measures = 30$. The described function was integrated with \gls{fdenser} on the model's validation step to measure the power used in the inference phase.

\begin{algorithm}[ht]
	\caption{Power Measure Algorithm}\label{alg:powerMeasure}
	\begin{algorithmic}
		\Require func, args, $n\_measures$
		
		\State $measures \gets \emptyset$

		\State $i \gets 1$
		
		\While{$i \leq n\_measures$}
		
		\State start(meter)
		
		\State $output \gets func(args)$
		
		\State stop(meter)
		
		\State $(energy$, $duration) \gets measure(meter)$
		
		\State $measure \gets energy / 1000 / duration$ \Comment{Convert mJ to W}
				
		\State $measures \gets measures \cup measure$
		
		\State $i \gets i + 1$	
		
		\EndWhile
		
		\State $mean\_power \gets mean(measures)$
		
		\State \Return ($output$, $mean\_power$)
	\end{algorithmic}
\end{algorithm}

It should be noted that ambient conditions of the server's location, such as temperature and humidity, were not considered, as well as other external variables, and no other processes used the GPU during the execution of these experiments.

\section{Results}
\label{sec:results}
This section compares the results from the baseline experiment and the experiment where our approaches were applied. The results show the mean accuracy and the mean power consumption, which are derived from the best individuals by fitness over 5 separate runs.

Since the results did not follow a normal distribution and the samples were independent, the Kruskal-Wallis non-parametric test was employed to determine if significant differences existed among the various approach groups. When significant differences were observed, the Mann-Whitney post-hoc test with Bonferroni correction was applied. We considered a significance level of $\alpha=0.05$ in all statistical tests.

\begin{figure}[ht]
    \centering
    \includesvg[width=\textwidth]{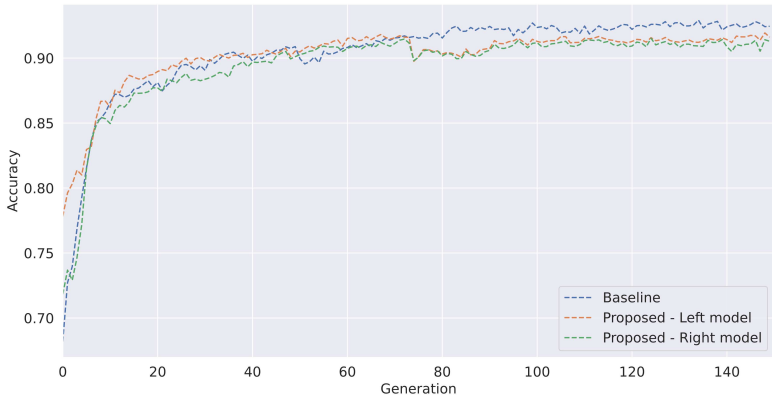}
    \caption{Evolution of the accuracy over 150 generations.} \label{fig:accuracy_comparison}
\end{figure}

Figure \ref{fig:accuracy_comparison} compares the accuracy obtained in the two experiments. The experiments present a similar accuracy until generation 70, where it becomes possible to observe a clear difference between them. The baseline experiment achieves a higher accuracy than the other experiment, and, relative to that experiment, it is visible that the smaller model obtains a marginally smaller accuracy than the larger one. Table \ref{table:comparison:mannwhitney_accuracy} provides statistical analysis, and Table \ref{table:comparison:stats_accuracy} showcases statistical values of the experiments. It is possible to see that, relative to the median values, the proposed method achieves inferior accuracy and that the smaller model obtains the worst accuracy.

\begin{table}[ht]
    \caption{Pair-wise comparison of used groups on accuracy metric, using Mann-Whitney U post-hoc test with Bonferroni correction with bold values denoting statistically significant differences.}
    \label{table:comparison:mannwhitney_accuracy}
    \centering
    \renewcommand{\arraystretch}{1.5}
  \begin{tabular}{cc|ccc}
  \cline{3-5}
                                          & \textbf{}      & \multicolumn{1}{c|}{\textbf{Baseline}} & \multicolumn{2}{c|}{\textbf{Proposed Method}}                 \\ \cline{2-5} 
  \multicolumn{1}{c|}{\textbf{}} &
    \textbf{Metric} &
    \multicolumn{1}{c|}{$Accuracy$} &
    \multicolumn{1}{c|}{$Accuracy_{left}$} &
    \multicolumn{1}{c|}{$Accuracy_{right}$} \\ \hline
  \multicolumn{1}{|c|}{\textbf{Baseline}} & $Accuracy$          & \cellcolor[HTML]{EFEFEF}               & \cellcolor[HTML]{EFEFEF} & \cellcolor[HTML]{EFEFEF} \\ \cline{1-3}
  \multicolumn{1}{|c|}{}                  & $Accuracy_{left}$ & \multicolumn{1}{c|}{\bm{$1.09 \times 10^{-4}$}}                 & \cellcolor[HTML]{EFEFEF} & \cellcolor[HTML]{EFEFEF} \\ \cline{2-4}
  \multicolumn{1}{|c|}{\multirow{-2}{*}{\textbf{\shortstack{Proposed\\Method}}}} &
    $Accuracy_{right}$ &
    \multicolumn{1}{c|}{\bm{$1.15 \times 10^{-7}$}} &
    \multicolumn{1}{c|}{\bm{$1.16 \times 10^{-4}$}} &
    \cellcolor[HTML]{EFEFEF} \\ \cline{1-4}
  \end{tabular}
\end{table}

\begin{table}[ht]
    \caption{Mean value, standard deviation, median and difference to baseline median of the accuracy of the experiments.}
    \label{table:comparison:stats_accuracy}
    \centering
    \renewcommand{\arraystretch}{1.5} 
    \begin{tabular}{|c|c|c|c|c|c|}
    \hline
    \textbf{Experiment}    & \textbf{Metric} & \textbf{Mean} & \textbf{SD}     & \textbf{Median} & \textbf{Diff. to Baseline} \\ \hline
    Baseline               & $Accuracy$           & 0.904             & 0.037 & 0.916 & \cellcolor[HTML]{EFEFEF}              \\ \hline
    \multirow{2}{*}{\shortstack{Proposed\\Method}} & $Accuracy_{left}$  & 0.902             & 0.024                  & 0.911 & $-0.005$              \\ \cline{2-6} 
                            & $Accuracy_{right}$ & 0.895             & 0.034                  & 0.907 & $-0.009$              \\ \hline
    \end{tabular}
\end{table}

\begin{figure}[ht]
    \centering
    \includesvg[width=\textwidth]{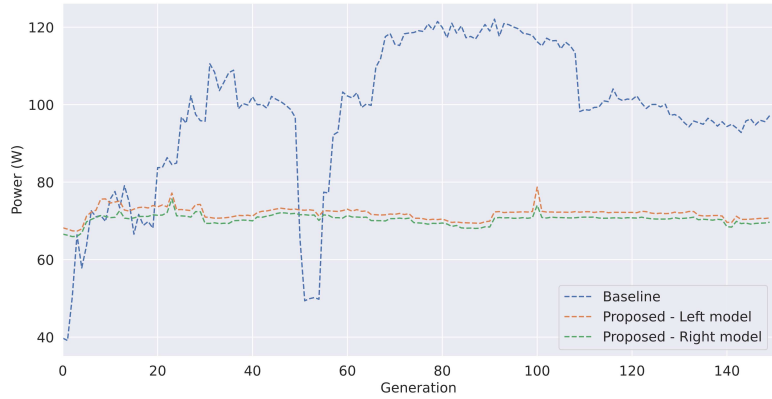}
    \caption{Evolution of the power consumption over 150 generations.} \label{fig:power_comparison}
\end{figure}

Figure \ref{fig:power_comparison} presents a comparison of the power consumption measured in the two experiments. The baseline predominantly has an increasing behaviour, which can be explained by the fact that the evolution is only being guided by accuracy, i.e., there are no incentives to favour models that consume less power. Contrarily, the proposed method obtained relatively stable results over the evolutionary process, with the smaller model presenting marginally lower results than its counterpart. Table \ref{table:comparison:mannwhitney_power} provides statistical analysis, and Table \ref{table:comparison:stats_power} showcases statistical values of the experiments. We can conclude that relative to the median values, the proposed method achieves inferior power consumption and that the smaller model is the most power-efficient.

\begin{table}[ht]
  \caption{Pair-wise comparison of used groups on power metric, using Mann-Whitney U post-hoc test with Bonferroni correction with bold values denoting statistically significant differences.}
  \label{table:comparison:mannwhitney_power}
  \centering
  \renewcommand{\arraystretch}{1.5}
  \begin{tabular}{cc|ccc}
  \cline{3-5}
                                          & \textbf{}      & \multicolumn{1}{c|}{\textbf{Baseline}} & \multicolumn{2}{c|}{\textbf{Proposed Method}}                 \\ \cline{2-5} 
  \multicolumn{1}{c|}{\textbf{}} &
    \textbf{Metric} &
    \multicolumn{1}{c|}{$Power$} &
    \multicolumn{1}{c|}{$Power_{left}$} &
    \multicolumn{1}{c|}{$Power_{right}$} \\ \hline
  \multicolumn{1}{|c|}{\textbf{Baseline}} & $Power$          & \cellcolor[HTML]{EFEFEF}               & \cellcolor[HTML]{EFEFEF} & \cellcolor[HTML]{EFEFEF} \\ \cline{1-3}
  \multicolumn{1}{|c|}{}                  & $Power_{left}$ & \multicolumn{1}{c|}{\bm{$2.72 \times 10^{-29}$}}                 & \cellcolor[HTML]{EFEFEF} & \cellcolor[HTML]{EFEFEF} \\ \cline{2-4}
  \multicolumn{1}{|c|}{\multirow{-2}{*}{\textbf{\shortstack{Proposed\\Method}}}} &
    $Power_{right}$ &
    \multicolumn{1}{c|}{\bm{$8.84 \times 10^{-32}$}} &
    \multicolumn{1}{c|}{\bm{$3.67 \times 10^{-19}$}} &
    \cellcolor[HTML]{EFEFEF} \\ \cline{1-4}
  \end{tabular}
\end{table}

\begin{table}[ht]
    \caption{Mean value, standard deviation, median and difference to baseline median of the experiments power consumption.}
    \label{table:comparison:stats_power}
    \centering
    \renewcommand{\arraystretch}{1.5} 
    \begin{tabular}{|c|c|c|c|c|c|}
    \hline
    \textbf{Experiment}    & \textbf{Metric} & \textbf{Mean} & \textbf{SD}     & \textbf{Median} & \textbf{Diff. to Baseline} \\ \hline
    Baseline               & $Power$           & 97.80 W             & 18.84 W & 99.89 W & \cellcolor[HTML]{EFEFEF}              \\ \hline
    \multirow{2}{*}{\shortstack{Proposed\\Method}} & $Power_{left}$  & 71.92 W             & 1.60 W                  & 72.20 W & $-27.69$ W              \\ \cline{2-6} 
                            & $Power_{right}$ & 70.40 W             & 1.30 W                  & 70.71 W & $-29.18$ W              \\ \hline
    \end{tabular}
\end{table}

\section{Conclusion}
\label{sec:conclusion}
In this work, we developed approaches integrated into \gls{fdenser}, which empower it to generate \gls{dnn} models with better power efficiency.

The most fundamental approach consists of measuring the power consumed by the \gls{gpu} on the inference phase of the \gls{dnn}. We use the measure provided by the \gls{gpu} to do this. Using this metric, we developed multi-objective fitness functions that steer the evolutionary process in a path that minimizes power consumption.

We created a process by which an additional output is added to a \gls{dnn} model and, after being trained, the model is split into two models -- a larger one which consists of all the layers and a smaller one composed of the layers up to the one where the additional output is connected to. This allows us to create models tuned for environments with fewer resources, such as smartphones, while creating more power-intensive models tuned for environments with more resources, such as servers. This is performed in one training, thus taking less time to develop the two models and saving energy in the process. No prior work has been identified that employs a similar approach.

We introduced a new mutation strategy to \gls{fdenser} that allows the reutilization of sets of layers -- modules -- according to the power consumption of the modules. We stochastically favour the reintroduction of modules in a model according to the inverse of the power they consume, thus incorporating power-efficient modules into a model.

The results obtained by our proposals show that we can reduce the power consumption of the \glspl{ann} without compromising their predictive performance, showing that it is possible to minimize power consumption while, at the same time, maximizing accuracy through the usage of \gls{ne} frameworks such as \gls{fdenser}. The best model found regarding power consumes 29.18 W (29.2\%) less whilst having a tiny decrease in performance (less than 1\%), proving that a small trade-off on accuracy can yield a considerable reduction in the power consumed by the model.

\subsection{Future Work}
\label{sec:conclusion:future_work}
We introduced novel approaches and performed a baseline experiment and an experiment where the mentioned strategies were applied. It could be valuable to explore other approaches and perform more experiments in the future.

To better understand the individual impact of each strategy on the efficiency of the models, it would be valuable to perform experiments with the employment of only one strategy at a time. It would also be interesting to vary the fitness functions (e.g., the weights used in them) and to vary evolutionary parameters such as the probabilities of the mutations.

One of the most important constraints of our work is \gls{gpu}-time due to the amount of operations required to train every model of each generation. To minimize the required time, it would be noteworthy to research how to employ training-less strategies in \gls{fdenser}, i.e., use strategies that estimate the accuracy of a model without training it \cite{chen2021neural,mellor2021neural}. Such strategies would allow us to perform more experiments in less time, saving energy in the design process.

\section*{Acknowledgments}
This work was supported by the Portuguese Recovery and Resilience Plan (PRR) through project C645008882-00000055, Center for Responsible AI, by the FCT, I.P./MCTES through national funds (PIDDAC), by Project No. 7059 - Neuraspace - AI fights Space Debris, reference C644877546-00000020, supported by the RRP - Recovery and Resilience Plan and the European Next Generation EU Funds, following Notice No. 02/C05-i01/2022, Component 5 - Capitalization and Business Innovation - Mobilizing Agendas for Business Innovation, and within the scope of CISUC R\&D Unit - UIDB/00326/2020.

\bibliographystyle{splncs04}
\bibliography{references}

\end{document}